\documentclass[final]{cvpr}

\usepackage{times}
\usepackage{epsfig}
\usepackage{graphicx}
\usepackage{amsmath}
\usepackage{amssymb}
\usepackage{color}
\usepackage[inline]{enumitem}

\usepackage[pagebackref=true,breaklinks=true,colorlinks,bookmarks=false]{hyperref}

\definecolor{greenblue}{RGB}{47,111,112}
\definecolor{darkred}{RGB}{158,28,28} 
\definecolor{darkgreen}{RGB}{28,158,28} 
\definecolor{orange}{RGB}{215,136,18}

\def\cvprPaperID{****} 
\def\confYear{CVPR 2021}

\begin{document}

\title{Semi-Supervised Disparity Estimation with Deep Feature Reconstruction}

\author{Julia Guerrero-Viu\thanks{Equal contribution}\qquad Sergio Izquierdo\footnotemark[1]\qquad Philipp Schr\"oppel\qquad Thomas Brox\\
University of Freiburg\\
{\tt\small guerrero,izquierd,schroepp,brox@cs.uni-freiburg.de}
}

\maketitle

\begin{abstract}
    Despite the success of deep learning in disparity estimation, the domain generalization gap remains an issue. We propose a semi-supervised pipeline that successfully adapts DispNet to a real-world domain by joint supervised training on labeled synthetic data and self-supervised training on unlabeled real data. Furthermore, accounting for the limitations of the widely-used photometric loss, we analyze the impact of deep feature reconstruction as a promising supervisory signal for disparity estimation.
\end{abstract}

\section{Introduction}
From 3D reconstruction to autonomous driving, many applications require depth estimates, which can be accurately obtained by predicting disparity from stereo images. 

Inspired by the matching cost and aggregation ideas from traditional stereo algorithms, recent works have designed successful deep learning architectures that are trained for disparity estimation on labeled data~\cite{mayer2016large, kendall2017end, guo2019gwc, cheng2020hierarchical, tankovich2020hitnet}. 
However, most of these works report only fine-tuned results for each individual dataset, neglecting generalization across domains ~\cite{kendall2017end, guo2019gwc, cheng2020hierarchical}.
Trying to improve this, and given the cost of acquiring ground truth data on real scenarios, some works have adopted a self-supervised approach~\cite{zhou2017unsupervised, yang2018segstereo, zhang2019dispsegnet, aleotti2020reversing}. They replace labels by a view reconstruction objective that maximizes consistency between the target image and the second image warped with the predicted disparity.
However, self-supervised methods are still less precise, especially on fine details and challenging areas, partially due to the limitations of this photometric consistency~\cite{jonschkowski2020unsupof}.
Aiming to overcome these limitations,~\cite{Zhan_2018_CVPR, shu2020feature} and~\cite{im2020unsupervised} have explored a reconstruction error based on deep features, for monocular depth and optical flow. 
Nevertheless, performance only improved marginally after combining it with a photometric loss, which shows the need for further analysis.

In this work, we present two contributions that address the described problems: \begin{enumerate*}[label=(\arabic*)] \item we propose a semi-supervised pipeline for disparity estimation that improves cross-domain generalization by exploiting cheap synthetic labels and unlabeled data from real scenarios, and \item we summarize the result of a thorough analysis of deep feature reconstruction (DFR) as consistency measure in self-supervised training, where we show examples of its potential and shed light on key reasons that limit its current effectiveness\end{enumerate*}.

\begin{figure}[t]
\begin{center}
\includegraphics[width=0.99\linewidth]{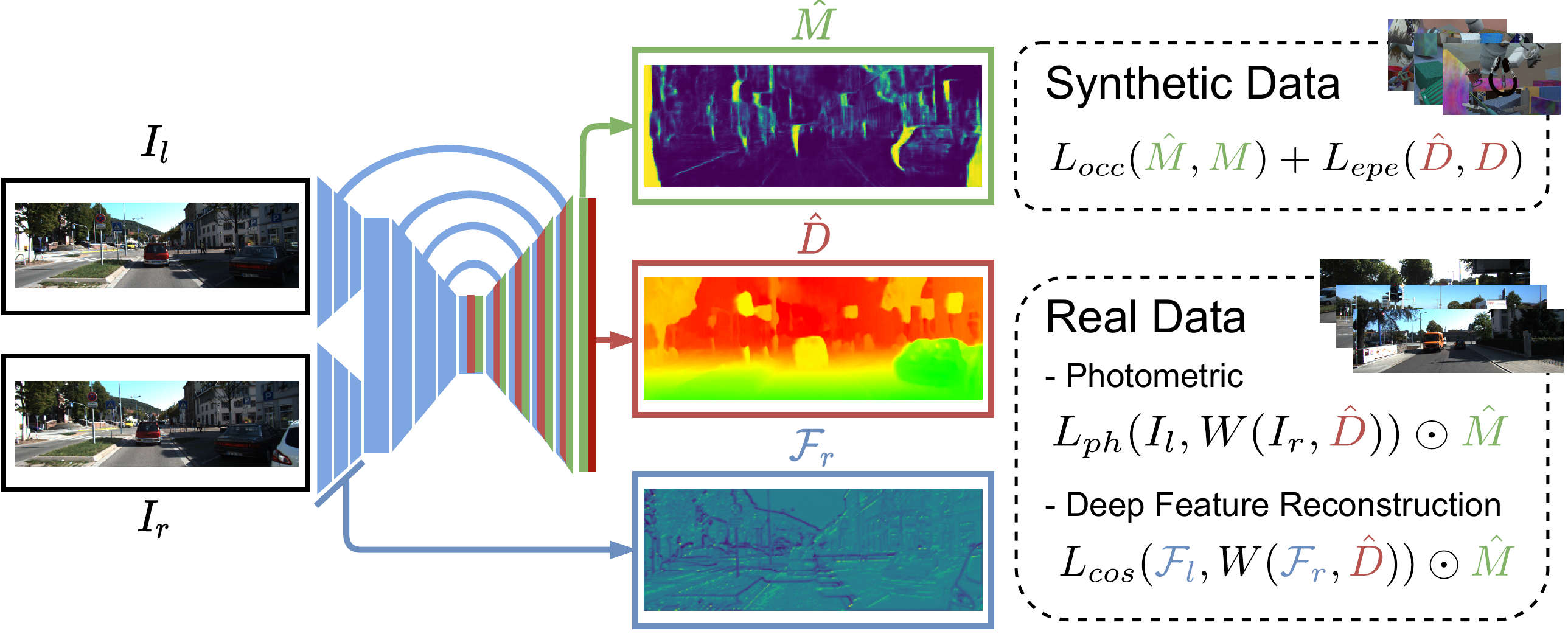}
\end{center}
   \caption{{\bf Semi-supervised training pipeline}: We use supervised training on synthetic data and self-supervised training on real data samples, either with photometric or deep feature reconstruction.} 
\label{fig:teaser}
\end{figure}

\begin{figure*}
\begin{center}
\includegraphics[width=0.95\linewidth]{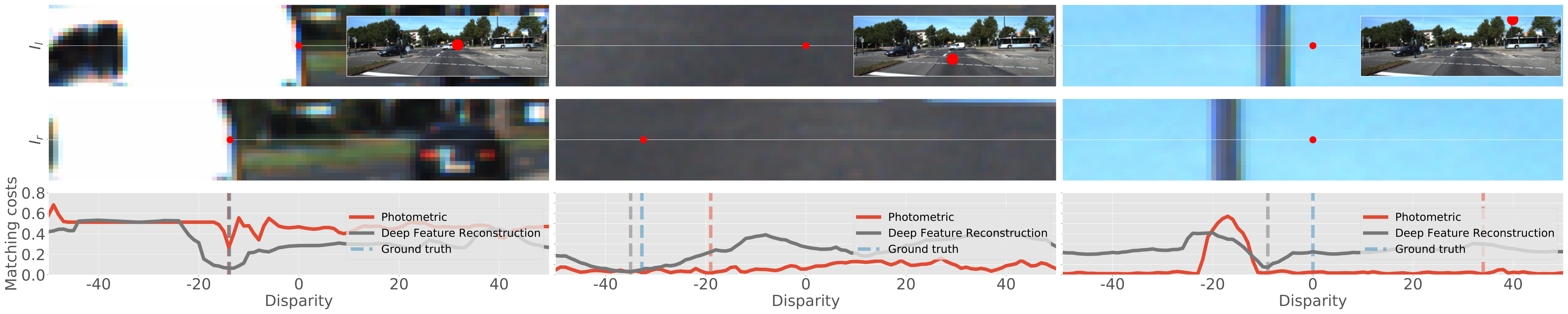}
\end{center}
   \caption{{\bf Matching costs curves of DFR and photometric on three different scenarios}. From top to bottom: zoom-in of the left image (target point in red); same zoom-in on the right image (matching ground truth in red); obtained matching costs at different disparities along the epipolar line, aligned with the right image (minimum of the curves and ground truth are marked with vertical dashed lines).}
\label{fig:responses}
\end{figure*}

\section{Semi-supervised pipeline}

Aiming to improve the domain adaptation, we propose a general semi-supervised pipeline for disparity estimation, depicted in Figure~\ref{fig:teaser}. We train a single end-to-end network by alternating batches of synthetic and real data. Synthetic batches, whose labels are already available at no extra cost, are used to train in a supervised manner. On the other hand, we train with real-data batches in a self-supervised way, using either the well-known photometric~\cite{godard2019digging} or deep feature reconstruction loss, as detailed in Section~\ref{section:dfr}. Additionally, following~\cite{ilg2018occ}, we train the network during supervised updates to predict occlusions. In real-data updates, we binarize the predicted occlusions and use them to mask out occluded areas in the self-supervised loss. Finally, we add synthetic occlusions~\cite{yang2019hierarchical, tankovich2020hitnet} to real-world data, which allows the network to learn about occlusions also from unlabeled samples.
In this work, we use the DispNet-C~\cite{mayer2016large} architecture, for its simplicity and efficiency, but our proposed pipeline is compatible with other disparity estimation networks.

\section{Deep feature reconstruction}\label{section:dfr}
We investigate on DFR for self-supervision by substituting the RGB images with features extracted from the first three layers of DispNet. Due to their higher-level representation, deep features should provide a more robust signal on challenging areas, like texture-less or non-Lambertian surfaces.
Given the predictions and feature maps at multiple scales, we resample each disparity map with nearest-neighbor interpolation to match all feature map sizes. We compute the dissimilarity between the left and the warped right feature maps using cosine distance, which resembles the computed dot product from the correlation layer.

\section{Experimental results}
We present results of our semi-supervised framework, using photometric loss (PH) and deep feature reconstruction (DFR).
We use FlyingThings3D (FT)~\cite{mayer2016large} for supervised updates, and KITTI Raw (K)~\cite{Geiger2013IJRR} for self-supervised ones, equally balancing their number of samples per epoch. 

\textbf{Semi-supervised learning with photometric consistency:}
In Table~\ref{tab:results}, we compare our results with two supervised baselines. On KITTI2015 (K15)~\cite{Menze2018JPRS}, the semi-supervised approach with photometric loss outperforms the supervised DispNet trained only on FT (-15.8\% EPE), showing the improved adaptation to real-world domains. We even achieve this while increasing the error on the synthetic FT domain only marginally (+4.7\% EPE), as opposed to the drastically increased error with supervised fine-tuning on K15 (+80.5\% EPE).
We also test the generalization to unseen environments using ETH3D~\cite{schoeps2017cvpr} and Middlebury at half resolution (MidH)~\cite{mid14}.
Remarkably, our semi-supervised approach generalizes also to these datasets, whereas fine-tuning on K15 only learns priors from KITTI but does not generalize to other real-world data. 

\setlength{\tabcolsep}{2.5pt}
\renewcommand{\arraystretch}{1.2}
\begin{table}
\scriptsize{}%
\centering
\begin{tabular}{@{\extracolsep{0pt}}lcc cccc}
&&& \multicolumn{4}{c}{Endpoint Error (EPE)}\\
\cline{4-7}
\multicolumn{1}{c}{Model} & train DS & time & FT & K15 & ETH3D & MidH\\
\hline
DispNet Supervised & \tiny{FT} & $0.04$ & $\mathbf{1.69}$ & $1.46$  & $0.92$ & $3.21$ \\
DispNet Supervised ft & \tiny{FT + K15} & $0.04$ & $3.05$ & (0.69) & $1.99$ & $3.79$ \\
\textbf{DispNet SemiSup. PH} & \tiny{FT + (K)} & $0.04$ & $1.77$ & $\mathbf{1.23}$ & $\mathbf{0.61}$ & $\mathbf{2.92}$ \\
\textbf{DispNet SemiSup. DFR} & \tiny{FT + (K)} & $0.04$ & $1.77$ & $1.32$ & $0.67$ & $2.94$ \\
\noalign{\smallskip}
\hline
\noalign{\smallskip}
GWCNet-gc~\cite{guo2019gwc} & \tiny{FT} & $0.32$ & $1.65$ & $2.35$ & $1.73$ & $5.08$\\
GWCNet-gc ft~\cite{guo2019gwc} & \tiny{FT + K12}  & $0.32$ & $5.63$ & $\mathbf{0.82}$ & $1.09$ & $5.41$ \\ 
LEA Stereo~\cite{cheng2020hierarchical}& \tiny{FT} & $0.30$ & $\mathbf{1.58}$ & $1.98$ & $0.87$ & $4.72$\\
Reversing PSMNet~\cite{aleotti2020reversing}& \tiny{(K)} & $0.41$ & $6.03$ & $1.01$ & $\mathbf{0.51}$ & $6.02$\\
\multicolumn{2}{c}{}&\multicolumn{1}{c}{}&\multicolumn{4}{c}{}\\
\end{tabular}
\caption{Results on FT 'cleanpass' test set and K15, ETH3D, MidH train sets. Train datasets are in brackets when no labels are used. We report all SOTA results by evaluating their publicly available models. 
Notice we do not filter disparities$>$192.
}
\label{tab:results}
\end{table}

\textbf{Deep feature reconstruction analysis:}
As shown in Table~\ref{tab:results}, we obtain inferior results with DFR than with its photometric counterpart. Instead of directly combining DFR with a photometric loss as in previous works, we conducted an in-depth analysis of the reasons that currently prevent the effectiveness of DFR. We identified the following issues: \begin{enumerate*}[label=(\arabic*)] \item higher sensitivity to occlusions, \item large dependence on the distance metric and resampling strategy, \item tainted information around disparity discontinuities due to convolutional aggregation, \item higher entropy on matching curves, and \item high gradient locality that complicates optimization\end{enumerate*}. We illustrate (3) and (4) on Figure~\ref{fig:responses}.
Despite DFR's clearly better response in texture-less areas (road), it fails near disparity discontinuities (sky-pole). Due to its higher-entropy curve, DFR is less precise on object boundaries (van).

\textbf{Comparison to the state of the art:}
Finally, we compare to state-of-the-art approaches, namely the supervised GWCNet~\cite{guo2019gwc} and LEAStereo~\cite{cheng2020hierarchical}, and the self-supervised Reversing PSMNet~\cite{aleotti2020reversing}. Results demonstrate the generalization problems of current supervised works. In contrast to this, our pipeline achieves reasonable results on FT and K15 and generalizes better across domains.

\section{Summary}
We have presented a novel semi-supervised pipeline and analyzed the influence of deep feature reconstruction for disparity estimation. Our results show improved generalization across domains, outperforming previous works in this setting. Based on our detailed study of DFR, we aim to exploit its potential in future work. 

\section*{Acknowledgment}
We acknowledge partial funding by 'la Caixa' Foundation (LCF/BQ/EU19/11710058), by the German Research Foundation (BR 3815/10-1) and by the German Federal Ministry for Economic Affairs and Energy within the project “KI Delta Learning – Development of methods and tools for the efficient expansion and transformation of existing AI modules of autonomous vehicles to new domains".

{\small
\bibliographystyle{ieee_fullname}
\bibliography{egbib}
}

\end{document}